\documentclass[sigconf]{acmart}

\usepackage{pifont}
\newcommand{\cmark}{\ding{51}}%
\newcommand{\xmark}{\ding{55}}%
\usepackage{latexsym}
\usepackage{amsmath,amsfonts}
\usepackage{amsthm}
\usepackage{algorithm}
\usepackage[noend]{algorithmic}
\usepackage{graphicx}
\usepackage{textcomp}
\usepackage{url}
\usepackage{xcolor}
\usepackage{colortbl}
\usepackage{bm}
\usepackage{epsfig}
\usepackage{epstopdf}
\usepackage{subfigure}
\usepackage{multirow}
\usepackage{array}
\usepackage{scalerel}
\usepackage{tikz}

\AtBeginDocument{%
  \providecommand\BibTeX{{%
    \normalfont B\kern-0.5em{\scshape i\kern-0.25em b}\kern-0.8em\TeX}}}

\setcopyright{acmcopyright}
\copyrightyear{2021}
\acmYear{2021}
\acmDOI{10.1145/1122445.1122456}

\acmConference[WWW '21]{WWW '21: The Web Conference}{April 19--23, 2021}{Ljubljana, Slovenia}
\acmBooktitle{WWW '21: The Web Conference,
  April 19--23, 2021, Ljubljana, Slovenia}
\acmPrice{15.00}
\acmISBN{978-1-4503-XXXX-X/18/06}

\begin{document}

\title{Slot Self-Attentive Dialogue State Tracking}

\author{Fanghua Ye}
\affiliation{
\institution{University College London, UK}
}
\email{fanghua.ye.19@ucl.ac.uk}

\author{Jarana Manotumruksa}
\affiliation{
\institution{University College London, UK}
}
\email{j.manotumruksa@ucl.ac.uk}

\author{Qiang Zhang}
\affiliation{
\institution{University College London, UK}
}
\email{qiang.zhang.16@ucl.ac.uk}

\author{Shenghui Li}
\affiliation{
\institution{Uppsala University, Sweden}
}
\email{shenghui.li@it.uu.se}

\author{Emine Yilmaz}
\affiliation{
\institution{University College London, UK}
}
\email{emine.yilmaz@ucl.ac.uk}


\renewcommand{\shortauthors}{Ye, et al.}


\begin{abstract}

  An indispensable component in task-oriented dialogue systems is the dialogue state tracker, which keeps track of users' intentions in the course of conversation. The typical approach towards this goal is to fill in multiple pre-defined slots that are essential to complete the task. Although various dialogue state tracking methods have been proposed in recent years, most of them predict the value of each slot separately and fail to consider the correlations among slots. In this paper, we propose a slot self-attention mechanism that can learn the slot correlations automatically. Specifically, a slot-token attention is first utilized to obtain slot-specific features from the dialogue context. Then a stacked slot self-attention is applied on these features to learn the correlations among slots. We conduct comprehensive experiments on two multi-domain task-oriented dialogue datasets, including MultiWOZ 2.0 and MultiWOZ 2.1. The experimental results demonstrate that our approach achieves state-of-the-art performance on both datasets, verifying the necessity and effectiveness of taking slot correlations into consideration.
  
\end{abstract}

\ccsdesc[500]{Computing methodologies~Natural language processing}
\ccsdesc[500]{Computing methodologies~Discourse, dialogue and pragmatics}
\ccsdesc[300]{Computing methodologies~Machine learning}

\keywords{dialogue state tracking, belief tracking, slot self-attention, task-oriented dialogue system}

\settopmatter{printfolios=true}
\maketitle

\section{Introduction} \label{sec:intro}
Task-oriented dialogue systems such as Apple Siri and Amazon Alexa work as virtual personal assistants. They can be leveraged to help users complete numerous daily tasks. A typical task-oriented dialogue system consists of four key components, i.e., natural language understanding (NLU), dialogue state tracking (DST), dialogue policy learning (DPL) and natural language generation (NLG) \cite{INR-074, chen2017survey}. Among them, DST aims at keeping track of users' intentions at each turn of the dialogue. Since DPL and NLU depend on the results of DST to select the next system action and generate the next system response, an accurate prediction of the dialogue state is crucial to enhance the overall performance of the dialogue system~\cite{kim-etal-2020-efficient, lee2019sumbt}. The typical dialogue state comprises a set of predefined slots and their corresponding values \cite{mrkvsic2017neural} (refer to Table~\ref{tab:toy} for an example). Therefore, the goal of DST is to predict the values of all slots at each  turn based on the dialogue context.

\begin{table}[]
\caption{An example dialogue with two domains. The value of slot ``{\itshape taxi-arriveby}'' should be inferred according to the value of slot ``{\itshape restaurant-book time}''. The value of slot ``{\itshape taxi-destination}'' is the same as that of slot ``{\itshape restaurant-name}''.}
\label{tab:toy}
\begin{tabular}{l}
\toprule
\textbf{Sys:} Hi, what can I do for you?                         \\
\textbf{User:} Please find me a \colorbox{red!50}{Chinese} restaurant.               \\
\textbf{State:} {\itshape restaurant-food=chinese}          \\ \midrule
\textbf{Sys:} \colorbox{red!50}{Charlie Chan} fits your criterion, can I book it for you?   \\
\textbf{User:} Yes, I need a table on \colorbox{red!50}{Monday} at \colorbox{red!50}{12:15}.              \\
\textbf{State:} {\itshape restaurant-food=chinese; restaurant-name=charlie chan} \\  \hspace{2.71em}{\itshape restaurant-book day=monday; restaurant-book time=12:15} \\ \midrule
\textbf{Sys:} Booking is successful. Is there anything else I can assist \\
\hspace{2.0em}you with today? \\
\textbf{User:} I also need a taxi to get me to \colorbox{orange!50}{the restaurant} \colorbox{blue!50}{on time}.      \\
\textbf{State:}  {\itshape restaurant-food=chinese; restaurant-name=charlie chan} \\  \hspace{2.71em}{\itshape restaurant-book day=monday; restaurant-book time=12:15} \\ \hspace{2.71em}{\itshape taxi-destination=charlie chan; taxi-arriveby=12:15} 
 \\ \bottomrule
\end{tabular}
\end{table}

DST has by far attracted much attention from both industry and academia, and numerous DST approaches have been proposed~\cite{williams2016dialog, henderson2015machine,dai2020survey, shan-etal-2020-contextual, kim2020somdst, heck2020trippy}. Although the state-of-the-art DST methods have achieved good performance, most of them predict the value of each slot separately, failing to consider the correlations among slots~\cite{hu2020sas,chen2020schema}. This can be problematic, as slots in a practical dialogue are unlikely to be entirely independent. Typically, some slots are highly correlated with each other, demonstrated by coreference and value sharing. Take the dialogue shown in Table~\ref{tab:toy} as an example. The value of slot ``{\itshape taxi-arriveby}'' is indicated by the slot ``{\itshape restaurant-book time}''. Thus, slot ``{\itshape taxi-arriveby}'' and slot ``{\itshape restaurant-book time}'' share the same value. The value of slot ``{\itshape taxi-destination}'' should also be taken from slot ``{\itshape restaurant-name}''. Furthermore, slot values can have a high co-occurrence probability. For example, the name of a restaurant should be highly relevant to the food type it serves.

In the literature, we notice that several DST approaches~\cite{chen2020schema, zhu2020efficient, hu2020sas} have tried to model the correlations among slots to a certain degree. However, these methods rely on huge human efforts and prior knowledge to determine whether two slots are related or not. As a consequence, they are severely deficient in scalability. Besides, they all leverage only the semantics of slot names to measure the relevance among slots and ignore the co-occurrences of slot values. Utilizing only the slot names is insufficient to capture the slot correlations completely and precisely. On one hand, the correlations among some slots may be overestimated, as slot values in a particular dialogue depend highly on the dialogue context. On the other hand, the correlations among some slots may be underestimated because their names have no apparent connections, even though their values have a high co-occurrence probability.

In this paper, we propose a new DST approach, named \textbf{S}lot self-a\textbf{T}tentive dialogue st\textbf{A}te t\textbf{R}acking (\textbf{STAR}), which takes both slot names and their corresponding values into account to model the slot correlations more precisely. 
Specifically, STAR first employs a slot-token attention module to extract slot-specific information for each slot from the dialogue context. It then utilizes a stacked slot self-attention module to learn the correlations among slots in a fully data-driven way. Hence, it does not ask for any human efforts or prior knowledge. The slot self-attention module also provides mutual guidance among slots and enhances the model's ability to deduce appropriate slot values from related slots. We conduct extensive experiments on both MultiWOZ 2.0 \cite{budzianowski2018multiwoz} and MultiWOZ 2.1 \cite{eric2019multiwoz} and show that STAR achieves better performance than existing methods that have taken slot correlations into consideration. STAR also outperforms other state-of-the-art DST methods\footnote{Code is available at \url{https://github.com/smartyfh/DST-STAR}}.

\section{Related Work}

DST is crucial to the success of a task-oriented dialogue system. Traditional statistical DST approaches rely on either the semantics extracted by the NLU module \cite{williams2007partially, thomson2010bayesian, williams2014web, wang2013simple} or some hand-crafted features and complex domain-specific lexicons \cite{henderson2014word, zilka2015incremental, mrkvsic2015multi, wen2017network, rastogi2018multi} to predict the dialogue state. These methods usually suffer from poor scalability and sub-optimal performance. They are also vulnerable to lexical and morphological variations \cite{lee2019sumbt, shan-etal-2020-contextual}.




Owing to the rise of deep learning, a neural DST model called neural belief tracking (NBT) has been proposed \cite{mrkvsic2017neural}. NBT employs convolutional filters over word embeddings in lieu of hand-crafted features to predict slot values. The performance of NBT is much better than previous DST methods. Inspired by this seminal work, a lot of neural DST approaches based on long short-term memory (LSTM) network \cite{zhong2018global, ren2018towards, nouri2018toward, ren2019scalable, rastogi2019scaling} and bidirectional gated recurrent unit (BiGRU) network \cite{rastogi2017scalable, wu2019transferable, mou2020multimodal, yang2020context, ouyang2020dialogue, hu2020sas} have been proposed for further improvements. These methods define DST as either a classification problem or a generation problem. Motivated by the advances in reading comprehension \cite{chen2018neural}, DST has been further formulated as a machine reading comprehension problem~\cite{gao2019dialog, ma2019end, gao2020machine, mou2020multimodal}. Other techniques such as pointer networks \cite{xu2018end} and reinforcement learning \cite{chen2020deep, huang2020meta, chen2020credit} have also been applied to DST. 


Recently, pre-training language models has gained much attention from both industry and academia, and a great variety of pre-trained language models such as BERT \cite{devlin2019bert} and GPT-2 \cite{radford2019language} have been released. Since the models are pre-trained on large corpora, they demonstrate strong abilities to produce good results when transferred to downstream tasks. In view of this, the research of DST has been shifted to building new models on top of these powerful pre-trained language models \cite{lee2019sumbt, shan-etal-2020-contextual, lin2020mintl, zhang2019find, kim2020somdst, gulyaev2020goal, wang2020dialogue, hosseini2020simple}. For example, SUMBT~\cite{lee2019sumbt} employs BERT to learn the relationships between slots and dialogue utterances through a slot-word attention mechanism. CHAN~\cite{shan-etal-2020-contextual} is built upon SUMBT via taking into account both slot-word attention and slot-turn attention. To better model dialogue behaviors during pre-training, TOD-BERT~\cite{wu2020tod} further pre-trains the original BERT model using several task-oriented dialogue datasets. SOM-DST \cite{kim2020somdst} considers the dialogue state as an explicit fixed-sized memory and selectively overwrites this memory to avoid predicting the dialogue state at each turn from scratch. TripPy \cite{heck2020trippy} uses three copy mechanisms to extract slot values. MinTL \cite{lin2020mintl} exploits T5 \cite{raffel2020exploring} and BART \cite{lewis2019bart} as the dialogue utterance encoder and jointly learns dialogue states and system responses. It also introduces Levenshtein belief spans to track dialogue states efficiently. NP-DST \cite{ham-etal-2020-end} and  SimpleTOD \cite{hosseini2020simple} adopt GPT-2 as the dialogue context encoder and  formulate DST as a language generation task.


All the methods mentioned above predict the value of each slot separately and ignore the correlations among slots.
We notice that several approaches \cite{chen2020schema, zhu2020efficient, hu2020sas} have tried to model the relevance among slots to a certain degree. Specifically, CSFN-DST~\cite{zhu2020efficient} and SST \cite{chen2020schema} construct a schema graph to capture the dependencies of different slots. However, the manually constructed schema graph is unlikely to reflect the correlations among slots completely. Besides, lots of prior knowledge is involved during the construction process. Therefore, CSFN-DST and SST are not scalable. SAS \cite{hu2020sas} calculates a slot similarity matrix to facilitate information flow among similar slots. The similarity matrix is computed based on either the cosine similarity or the K-means clustering results of slot names. However, when computing the similarity matrix, SAS involves several hyperparameters, which are hard to set. SAS also fixes the similarity coefficient at 1 if two slots are considered to be relevant. This is obviously impractical. Except for the model-specific drawbacks of CSFN-DST, SST and SAS, they also share a common limitation: they all measure the slot correlations using only the slot names. This may overlook or overrate the dependencies of some slots. Our method utilizes both slot names and their corresponding values to model slot correlations more precisely.

\section{Preliminaries}

In this section, we first provide the formal definition of DST and then conduct a simple data analysis to show the high correlations among slots in practical dialogues.

\subsection{Problem Statement}

The goal of DST is to extract a set of slot value pairs from the system response and user utterance at each turn of the conversation. The combination of these slot value pairs forms a dialogue state, which keeps track of the complete intentions or requirements that have been informed by the user to the system. Formally, let $\mathcal{X} = \{(R_1, U_1), (R_2, U_2), \dots, (R_T, U_T)\}$ denote a conversation of $T$ turns, where $R_t$ and $U_t$ represent the system response and user utterance at turn $t$, respectively. Suppose that we have a set of $J$ predefined slots $\mathcal{S}=\{S_1, S_2, \dots, S_J\}$ with $S_j$ being the $j$-th slot, then the dialogue state at turn $t$ is defined as $\smash{\mathcal{B}_t = \{(S_j, V_j^t) | 1 \leq j \leq J \}}$, where $\smash{V_j^t \in \mathcal{V}_j}$ denotes the corresponding value of slot $S_j$. $\mathcal{V}_j$ is the value space of slot $S_j$. Putting the value spaces of all slots together, we construct an ontology $\mathbb{O} = \{(S_1, \mathcal{V}_1), (S_2, \mathcal{V}_2), \dots, (S_J, \mathcal{V}_J)\}$. Based on the dialogue context $\mathcal{X}$ and the ontology $\mathbb{O}$, the task of DST is defined as learning a dialogue state tracker $\mathcal{F}: \mathcal{X} \times \mathbb{O} \rightarrow \mathcal{B}$ that can efficiently capture the user's intentions in the course of conversation. According to this definition, we can see that DST is a relatively challenging problem, as it is needed to predict the values of multiple slots at each turn. Besides, the value spaces of some slots may be large, that is, there may be a large number of candidate values for some slots. This phenomenon makes the prediction of dialogue states even more challenging.

It is worth mentioning that in this paper we use the term ``slot'' to refer to the concatenation of the domain name and the slot name so as to include both domain and  slot information. For example, we use ``{\itshape restaurant-pricerange}'' rather than ``{\itshape pricerange}'' to represent the ``{\itshape pricerange}'' slot in the ``{\itshape restaurant}'' domain. This format is useful, especially when a conversation involves multiple domains. It has also been widely adopted by previous works \cite{lee2019sumbt, shan-etal-2020-contextual, kim2020somdst, hu2020sas}.

\begin{figure}[t!]
	\centering
	\subfigure[{Restaurant Area}]{
		\includegraphics[width=0.4765\columnwidth]{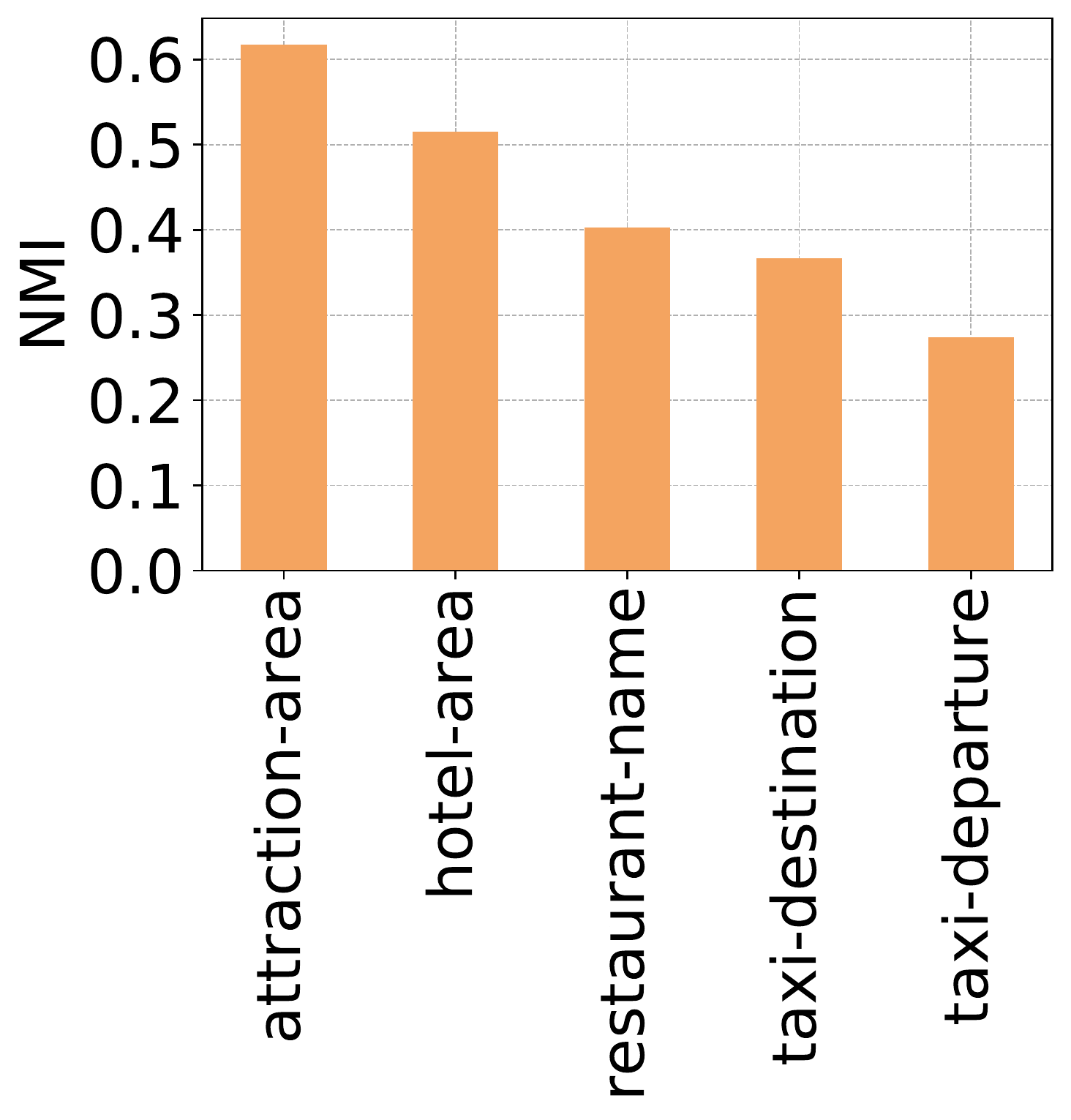}
	} \hfill
	\subfigure[{Taxi Destination}]{
		\includegraphics[width=0.4766\columnwidth]{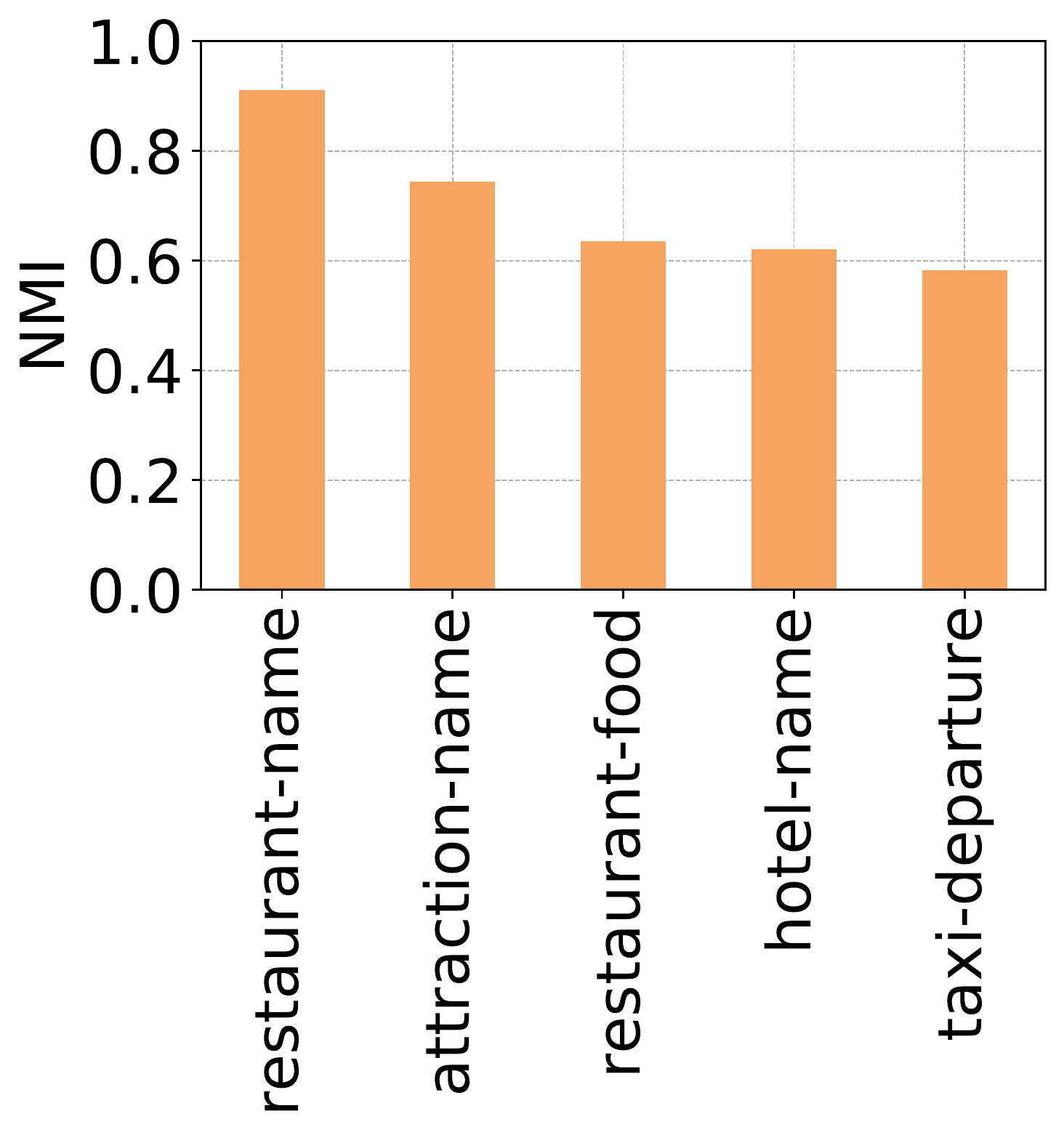}
	}
	\vspace*{-0.2cm}
	
	\caption{The top-5 most correlated slots of slot ``{\itshape restaurant-area}'' and slot ``{\itshape taxi-destination}'' analyzed on MultiWOZ 2.1. The slot itself is not counted as a relevant one.}
	\label{fig:corr}
\end{figure}

\subsection{Data Analysis}

To intuitively verify the strong correlations among slots in practical dialogues, we conduct a simple data analysis on MultiWOZ 2.1 \cite{eric2019multiwoz}, which is a multi-domain task-oriented dialogue dataset. Specifically, we treat every slot pair as two different partitions of the dataset. For each partition, the corresponding slot values are regarded as the cluster labels. Then we calculate the normalized mutual information (NMI) score between the two partitions. Note that we adopt NMI as the measurement of slot  correlations, as mutual information can describe more general dependency relationships beyond linear dependence\footnote{\url{https://en.wikipedia.org/wiki/Mutual_information}}. We illustrate the top-5 most relevant slots of slot ``{\itshape restaurant-area}'' and slot ``{\itshape taxi-destination}'' in Figure~\ref{fig:corr}. Other slots show similar patterns. From Figure~\ref{fig:corr}, we observe that slot ``{\itshape restaurant-area}'' and slot ``{\itshape taxi-destination}'' are indeed highly correlated with some other slots. The relevant slots are not only within the same domain but also across different domains. For example, slot ``{\itshape taxi-destination}'' correlates highly with slot ``{\itshape restaurant-food}'', even though their names have no apparent connections. This observation consolidates our motivation that it is necessary to take into account both slot names and their values.


\section{STAR: Slot Self-Attentive DST}

\begin{figure*}[t]
  \centering
  \includegraphics[width=0.83\textwidth]{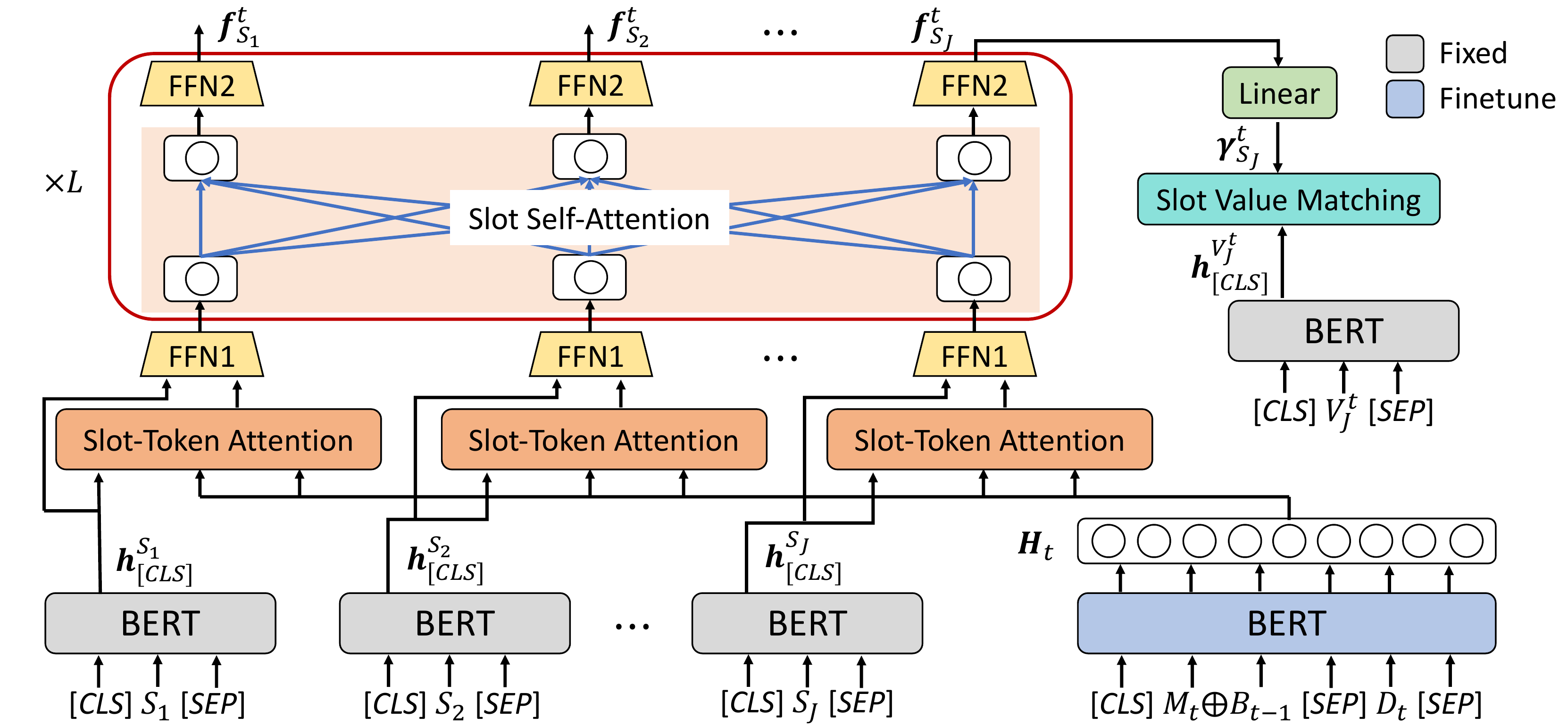}
  \caption{The architecture of our approach STAR. A fine-tuning BERT is used to encode dialogue contexts, another fixed BERT is utilized to generate aggregated vector representations for slots and values. For simplicity, layer normalization and residual connection are omitted, and only the value matching of slot $S_J$ is included. Both FFN1 and FFN2 are feed-forward networks.}
  \label{fig:architecture}
\end{figure*}

In this section, we describe our proposed slot self-attentive DST model {\itshape STAR} in detail. The overall architecture of STAR is illustrated in Figure~\ref{fig:architecture}, which is composed of a BERT-based context encoder module, a slot-token attention module, a stacked slot self-attention module and a slot value matching module.

\subsection{Context Encoder}

Recently, many pre-trained language models such as BERT \cite{devlin2019bert} and GPT-2 \cite{radford2019language} have shown strong abilities to produce good results when transferred to downstream tasks. In view of this, we employ BERT as the context encoder to obtain semantic vector representations of dialogue contexts, slots and values. BERT is a deep bidirectional language representation learning model rooted in Transformer encoders \cite{vaswani2017attention}. It can generate token-specific vector representations for each token in the input sentence as well as an aggregated vector  representation of the whole sentence. Therefore, we exploit BERT to generate token-specific vector representations for dialogue contexts and aggregated vector representations for both slots and values.

\subsubsection{Dialogue Context Encoder}
The dialogue utterances at turn $t$ are represented as $D_t = R_t \oplus U_t$, where $\oplus$ is the operation of sequence concatenation. The dialogue history of turn $t$ is denoted as $M_t = D_1 \oplus D_2 \oplus \cdots \oplus D_{t-1}$. Then, the entire dialogue context of turn $t$ is defined as:
\begin{equation}
    X_t = [CLS] \oplus M_t \oplus [SEP] \oplus D_t \oplus [SEP],
\end{equation}
where $[CLS]$ and $[SEP]$ are two special tokens introduced by BERT. The $[CLS]$ token is leveraged to aggregate all token-specific representations and the $[SEP]$ token is utilized to mark the end of a sentence. Since the maximum input length of BERT is restricted to 512 \cite{devlin2019bert}, we must truncate $X_t$ if it is too long. The straightforward way is to cut off the early dialogue history and reserve the most recent one in $M_t$. However, this operation may throw away some key information. To reduce information loss, we use the previous dialogue state as input as well, which is expected to keep all the slot-related history information. The dialogue state of previous turn is represented by $B_{t-1} = \bigoplus_{(S_j, V_j^{t-1}) \in \mathcal{B}_{t-1}, V_j^{t-1} \neq \verb|none|} S_j \oplus V_j^{t-1}$. Note that in $B_{t-1}$ we only include the slots that have been mentioned before (i.e., only  non-\verb|none| slots are considered). By treating $B_{t-1}$ as part of the dialogue history, the entire dialogue context of turn $t$ is finally denoted as\footnote{Although the previous dialogue state $B_{t-1}$ can serve as an explicit, compact and informative representation of the dialogue history, we find that it is still useful to take the dialogue history $M_t$ as part of the input.}: 
\begin{equation}
    X_t = [CLS] \oplus M_t \oplus B_{t-1} \oplus [SEP] \oplus D_t \oplus [SEP].
\end{equation}

Let $|X_t|$ be the number of tokens in $X_t$. Our first goal is to generate a contextual $d$-dimensional vector representation for each token in $X_t$. Let $\bm{h}_j^t \in \mathbb{R}^d$ denote the vector representation of the $j$-th token and $\smash{\bm{H}_t = [\bm{h}_1^t, \bm{h}_2^t, \dots, \bm{h}_{|X_t|}^t] \in \mathbb{R}^{d \times |X_t|}}$ the matrix form of all tokens' representations. We simply feed $X_t$ to BERT to obtain $\bm{H}_t$. Hence, we have:
\begin{equation} \label{eqn:input}
    \bm{H}_t = BERT_{finetune} (X_t).
\end{equation}
Note that BERT in Eq.~\eqref{eqn:input} will be fine-tuned during training.

\subsubsection{Slot and Value Encoder}
Following previous works \cite{lee2019sumbt, shan-etal-2020-contextual}, we use another BERT to encode slots and their candidate values. Unlike dialogue contexts, we need to generate aggregated vector representations for slots and values. To achieve this goal, we use the vector representation corresponding to the special token $[CLS]$ to represent the aggregated representation of the whole input sequence. As thus, for any slot $S_j \in \mathcal{S} (1 \leq j \leq J)$ and any value $V_j^t \in \mathcal{V}_j$, we have:
\begin{equation} \label{eq:slotandlabel}
    \begin{gathered}
        \bm{h}_{[CLS]}^{S_j} = BERT_{fixed}([CLS] \oplus S_j \oplus [SEP]),\\
        \bm{h}_{[CLS]}^{V_j^t} = BERT_{fixed}([CLS] \oplus V_j^t \oplus [SEP]),
    \end{gathered}
\end{equation}
where $BERT_{fixed}$ means that the pre-trained BERT without fine-tuning is adopted. Fixing the weights of BERT when encoding slots and values is beneficial. Firstly, the slot and value representations can be computed off-line, which reduces the model size of our approach. Secondly, since our model relies on the value representations to score each candidate value of a given slot, fixing the representations of values can reduce the difficulty of choosing the best candidate value.


\subsection{Slot-Token Attention}

Since there are multiple slots to be predicted at each turn $t$ from the same dialogue context $X_t$, it is necessary to extract slot-specific information for each slot $S_j (1 \leq j \leq J)$. Our model employs a multi-head attention mechanism \cite{vaswani2017attention} to retrieve the relevant information corresponding to each slot $S_j$. 

\subsubsection{Multi-Head Attention}

For self-consistency, we provide a brief description of the multi-head attention mechanism. Assume that we are provided with a query matrix $\bm{Q} = [\bm{q}_1, \bm{q}_2, \dots,  \bm{q}_{|Q|}] \in \mathbb{R}^{d_{model} \times |Q|}$, a key matrix $\smash{\bm{K}= [\bm{k}_1, \bm{k}_2, \dots, \bm{k}_{|K|}] \in \mathbb{R}^{d_{model} \times |K|}}$ and a value matrix $\smash{\bm{Z} = [\bm{z}_1, \bm{z}_2, \dots,  \bm{z}_{|K|}] \in \mathbb{R}^{d'_{model} \times |K|}}$. There are $|Q|$ query vectors, $|K|$ key vectors and $|K|$ value vectors, respectively. Note that the query vectors and the key vectors share the same dimensionality. For each query vector $\bm{q}_i (1 \leq i \leq |Q|)$, the attention vector $\bm{a}_i$ over $\bm{K}$ and $\bm{Z}$ with $N$ heads is calculated as follows:
\begin{equation*}
\begin{gathered}
    e_{ij}^n = \frac{(\bm{W}_Q^n \bm{q}_i)^T (\bm{W}_K^n \bm{k}_j)}{\sqrt{d_{model} / N}}, \quad 
    \tau_{ij}^n = \frac{exp(e_{ij}^n)}{\sum_{l=1}^{|K|} exp(e_{il}^n)}, \\
    \bm{a}_i^n = \sum_{j=1}^{|K|} \tau_{ij}^n (\bm{W}_Z^n \bm{z}_j), \quad
    \bm{a}_i =  \bm{W}_O Concat(\bm{a}_i^1, \bm{a}_i^2, \dots, \bm{a}_i^N), 
\end{gathered}
\end{equation*}
where $1 \leq n \leq N$, $\bm{a}_i \in \mathbb{R}^{d'_{model}}$,  $\bm{W}_Q^n \in \mathbb{R}^{(d_{model} / N) \times d_{model}}$, $\bm{W}_K^n \in \mathbb{R}^{(d_{model} / N) \times d_{model}}$, $\bm{W}_Z^n \in \mathbb{R}^{(d'_{model} / N) \times d'_{model}}$ and $\bm{W}_O \in \mathbb{R}^{d'_{model} \times d'_{model}}$. Putting all the attention vectors together, we obtain the attention matrix $\bm{A} = [\bm{a}_1, \bm{a}_2, \dots, \bm{a}_{|Q|}] \in \mathbb{R}^{d'_{model} \times |Q|}$. The entire process is formulated as:  \begin{equation*}
    \bm{A} = MultiHead(\bm{Q}, \bm{K}, \bm{Z}).
\end{equation*}

\subsubsection{Multi-Head Slot-Token Attention}
Our model adopts the multi-head attention mechanism to calculate a $d$-dimensional vector for each slot $S_j$ as the slot-specific information. More concretely, the slot representation $\smash{\bm{h}_{[CLS]}^{S_j}}$ is treated as the query vector, and the dialogue context representation $\bm{H}_t$ is regarded as both the key matrix and the value matrix. Consequently, the token-level relevance between slot $S_j$ and dialogue context $X_t$ is summarized as:
\begin{equation}
    \bm{r}_{S_j}^t = MultiHead(\bm{h}_{[CLS]}^{S_j}, \bm{H}_t, \bm{H}_t), 
\end{equation}
where $\smash{\bm{r}_{S_j}^t \in \mathbb{R}^d}$. Considering that $\smash{\bm{r}_{S_j}^t}$ only contains the value information of slot $S_j$, we concatenate $\smash{\bm{r}_{S_j}^t}$ and $\smash{\bm{h}_{[CLS]}^{S_j}}$ to retain its name information. This merged vector is further transformed by a feed-forward neural network as below:
\begin{equation}
    \bm{c}_{S_j}^t = \bm{W}_2^r ReLU(\bm{W}_1^r Concat(\bm{h}_{[CLS]}^{S_j}, \bm{r}_{S_j}^t) + \bm{b}_1^r) + \bm{b}_2^r, 
\end{equation}
where $\bm{W}_1^r \in \mathbb{R}^{d \times 2d}$, $\bm{W}_2^r \in \mathbb{R}^{d \times d}$ and $\bm{b}_1^r, \bm{b}_2^r, \bm{c}_{S_j}^t \in \mathbb{R}^{d}$.

\subsection{Slot Self-Attention}

Albeit the slot-token attention is expected to retrieve slot-specific information for all slots, it may fail to capture the valid information of some slots due to the various expressing forms in natural conversations (e.g., coreference, synonymity and rephrasing). In addition, the slot-specific vector $\smash{\bm{c}_{S_j}^t}$ of each slot $S_j$ is computed separately. The correlations among slots are ignored. As a result, once the vector $\smash{\bm{c}_{S_j}^t}$ doesn't capture the relevant information of slot $S_j$ properly, the model has no chance to deduce the right value for slot $S_j$. To alleviate this problem, we propose exploiting the slot self-attention mechanism to rectify each slot-specific vector based on the vectors corresponding to all slots. This mechanism should be rational because of the high correlations among slots. Therefore, our model is expected to provide mutual guidance among slots and learn the slot correlations automatically.

The slot self-attention is also a multi-head attention. Specifically, this module is composed of $L$ identical layers and each layer has two sub-layers. The first sub-layer is the slot self-attention layer. The second sub-layer is a feed-forward network (FFN) with two fully connected layers and a ReLU activation in between. Each sub-layer precedes its main functionality with layer normalization \cite{ba2016layer} and follows it with a residual connection \cite{he2016deep}.

Let $\smash{\bm{C}_t = [\bm{c}_{S_1}^t, \bm{c}_{S_2}^t, \dots, \bm{c}_{S_J}^t] \in \mathbb{R}^{d \times J}}$ denote the matrix representation of all slot-specific vectors and let  $\bm{F}_t^1 = \bm{C}_t$.  Then, for the $l$-th slot self-attention sub-layer ($1 \leq l \leq L$), we have:
\begin{equation}
\begin{gathered}
    \tilde{\bm{F}}_t^l = LayerNorm(\bm{F}_t^{l}), \\
    \bm{G}_t^l = MultiHead(\tilde{\bm{F}}_t^l, \tilde{\bm{F}}_t^l, \tilde{\bm{F}}_t^l) + \tilde{\bm{F}}_t^l.
\end{gathered}
\end{equation}
In the slot self-attention sub-layer, $\tilde{\bm{F}}_t^l$ serves as the key matrix, the value matrix, and also the query matrix. For the $l$-th feed-forward sub-layer, we have:
\begin{equation}
\begin{gathered}
    \tilde{\bm{G}}_t^l = LayerNorm(\bm{G}_t^{l}), \\
    \bm{F}_t^{l+1} = FFN(\tilde{\bm{G}}_t^l) + \tilde{\bm{G}}_t^l,
\end{gathered}
\end{equation}
where the function $FFN(\cdot)$ is parameterized by $\bm{W}_1, \bm{W}_2 \in \mathbb{R}^{d \times d}$ and $\bm{b}_1, \bm{b}_2 \in \mathbb{R}^d$, i.e., $FFN(\bm{y}) = \bm{W}_2 ReLU(\bm{W}_1\bm{y} + \bm{b}_1) + \bm{b}_2$.

The final slot-specific vectors are contained in the output of the last layer, i.e., $\bm{F}_t^{L+1}$. Let $\smash{\bm{F}_t^{L+1} = [\bm{f}_{S_1}^t, \bm{f}_{S_2}^t, \dots, \bm{f}_{S_J}^t]}$, where $\smash{\bm{f}_{S_j}^t \in \mathbb{R}^d}$ is the $j$-th column of $\bm{F}_t^{L+1}$. $\smash{\bm{f}_{S_j}^t}$ is taken as the final slot-specific vector of slot $S_j$, which is expected to be close to the semantic vector representation of the groundtruth value of slot $S_j$ at turn $t$. Since the output of BERT is normalized by layer normalization, we also feed $\smash{\bm{f}_{S_j}^t}$ to a layer normalization layer, which is preceded by a linear transformation layer as follows:
\begin{equation}
    \bm{\gamma}_{S_j}^t = LayerNorm(Linear(\bm{f}_{S_j}^t)),
\end{equation}
where $\smash{\bm{\gamma}_{S_j}^t \in \mathbb{R}^d}$.

\subsection{Slot Value Matching}

To predict the value of each slot $S_j (1 \leq j \leq J)$, we compute the distance between $\smash{\bm{\gamma}_{S_j}^t}$ and the semantic vector representation of each value $V'_j \in \mathcal{V}_j$, where $\mathcal{V}_j$ denotes the value space of slot $S_j$. Then the value with the smallest distance is chosen as the prediction of slot $S_j$. We adopt $\ell_2$ norm as the distance metric.

During the training phase, we calculate the probability of the groundtruth value $V_j^t$ of slot $S_j$ at turn $t$ as:
\begin{equation}
    p(V_j^t | X_t, S_j) = \frac{exp\left(-\Vert \bm{\gamma}_{S_j}^t - \bm{h}_{[CLS]}^{V_j^t} \Vert_2\right)}{\sum_{V'_j \in \mathcal{V}_j} exp\left(-\Vert \bm{\gamma}_{S_j}^t - \bm{h}_{[CLS]}^{V'_j} \Vert_2\right)}.
\end{equation}
Our model is trained to maximize the joint probability of all slots, i.e., $\smash{\Pi_{j=1}^J p(V_j^t | X_t, S_j)}$. For this purpose, the loss function at each turn $t$ is defined as the sum of the negative log-likelihood:
\begin{equation}
    \mathcal{L}_t = \sum_{j=1}^J - \log\left(p(V_j^t | X_t, S_j)\right).
\end{equation}

\section{Experimental Setup}

\subsection{Datasets}

\begin{table}[h]
\caption{Data statistics of MultiWOZ 2.0 \& 2.1. The bottom row summarizes the number of dialogues in each domain.}
\label{tab:datastat}
\footnotesize
\setlength{\tabcolsep}{1.3mm}{
\begin{tabular}{lccccc}
\toprule
Domain            & Attraction & Hotel & Restaurant & Taxi & Train \\ \midrule
Slot &
  \begin{tabular}[c]{@{}c@{}}area\\ name\\ type\end{tabular} &
  \begin{tabular}[c]{@{}c@{}}area\\ book day\\ book people\\ book stay\\ internet\\ name\\ parking\\ pricerange\\ stars\\ type\end{tabular} &
  \begin{tabular}[c]{@{}c@{}}area\\ book day\\ book people\\ book time\\ food\\ name\\ pricerange\end{tabular} &
  \begin{tabular}[c]{@{}c@{}}arriveby\\ departure\\ destination\\ leaveat\end{tabular} &
  \begin{tabular}[c]{@{}c@{}}arriveby\\ book people\\ day\\ departure\\ destination\\ leaveat\end{tabular} \\ \midrule
Train       & 2717       & 3381  & 3813       & 1654 & 3103  \\
Validation & 401        & 416   & 438        & 207  & 484   \\
test        & 395        & 394   & 437        & 195  & 494   \\ \bottomrule
\end{tabular}}
\end{table}

We evaluate our approach on MultiWOZ 2.0 \cite{budzianowski2018multiwoz} and MultiWOZ 2.1 \cite{eric2019multiwoz}, which are two of the largest publicly available task-oriented dialogue datasets. MultiWOZ 2.0 consists of 10,348 multi-turn dialogues, spanning over 7 domains $\{${\itshape attraction}, {\itshape hotel}, {\itshape restaurant}, {\itshape taxi}, {\itshape train}, {\itshape hospital}, {\itshape police}$\}$. Each domain has multiple predefined slots and there are 35 domain slot pairs in total. MultiWOZ 2.1 is a refined version of MultiWOZ 2.0. According to \cite{eric2019multiwoz}, about $32\%$ of the state annotations have been corrected in MultiWOZ 2.1. Since {\itshape hospital} and {\itshape police} are not included in the validation set and test set, following previous works \cite{wu2019transferable, shan-etal-2020-contextual, lee2019sumbt, kim2020somdst, zhu2020efficient}, we use only the remaining 5 domains in the experiments. The resulting datasets contain 17 distinct slots and 30 domain slot pairs. The detailed statistics are summarized in Table~\ref{tab:datastat}.

We follow similar data preprocessing procedures as \cite{wu2019transferable} to preprocess both MultiWOZ 2.0 and MultiWOZ 2.1. And we create the ontology by incorporating all the slot values that appear in the datasets. We notice that several works \cite{shan-etal-2020-contextual, lee2019sumbt} exploit the original ontology provided by MultiWOZ 2.0 and MultiWOZ 2.1 to preprocess the datasets in their experiments. However, the original ontology is incomplete. If a slot value is out of the ontology, this value is ignored directly in \cite{shan-etal-2020-contextual, lee2019sumbt}, which is impractical and leads to unreasonably high performance.

\subsection{Comparison Methods}

We compare our model STAR with the following state-of-the-art DST approaches: 
\begin{itemize}
    \item \textbf{SST:} SST \cite{chen2020schema} first constructs a schema graph and then utilizes a graph attention network (GAT) \cite{velivckovic2017graph} to fuse information from dialogue utterances and the schema graph. 
    
    \item \textbf{SAS:} SAS \cite{hu2020sas} calculates a binary slot similarity matrix to control information flow among similar slots. The similarity matrix is computed via either a fixed combination method or a K-means sharing method. 
    
    \item \textbf{CREDIT-RL:} CREDIT-RL \cite{chen2020credit} employs a structured representation to represent dialogue states and casts DST as a sequence generation problem. It also uses a reinforcement loss to fine-tune the model. 
    
    \item \textbf{STARC:} STARC \cite{gao2020machine} reformulates DST as a machine reading comprehension problem and adopts several reading comprehension datasets as auxiliary information to train the model.
    
    \item \textbf{CSFN-DST:} Similar to SST, CSFN-DST \cite{zhu2020efficient} also constructs a schema graph to model the dependencies among slots. However, CSFN-DST utilizes BERT \cite{devlin2019bert} rather than GAT to encode  dialogue utterances.
    
    \item \textbf{SOM-DST:} SOM-DST \cite{kim2020somdst} regards the dialogue state as an explicit fixed-sized memory and proposes selectively overwriting this memory at each turn. 
    
    \item \textbf{CHAN:} CHAN \cite{shan-etal-2020-contextual} proposes a slot-turn attention mechanism to make full use of the dialogue history. It also designs an adaptive objective to alleviate the data imbalance issue. 
    
    \item \textbf{TripPy:} TripPy \cite{heck2020trippy} leverages three copy mechanisms to extract slot values from user utterances, system inform memory and previous dialogue states.
    
    \item \textbf{NP-DST:} NP-DST \cite{ham-etal-2020-end} transforms DST into a language generation problem and adopts GPT-2 \cite{radford2019language} as both the dialogue context encoder and the sequence generator. 
    
    \item \textbf{SimpleTOD:} SimpleTOD \cite{hosseini2020simple} is also based on GPT-2. Its model architecture is similar to NP-DST. 
    
    \item \textbf{MinTL:} MinTL \cite{lin2020mintl} is an effective transfer learning framework for task-oriented dialogue systems. It introduces Levenshtein belief span to track dialogue states. MinTL uses both T5 \cite{raffel2020exploring} and BART \cite{lewis2019bart} as pre-trained backbones. We name them \textbf{MinTL-T5} and \textbf{MinTL-BART} for distinction. 
\end{itemize}

\subsection{Evaluation Metric}

We adopt joint goal accuracy \cite{nouri2018toward} as the evaluation metric. Joint goal accuracy is defined as the ratio of dialogue turns for which the value of each slot is correctly predicted. If a slot has not been mentioned yet, its groundtruth value is set to \verb|none|. All the \verb|none| slots also need to be predicted. Joint goal accuracy is a relatively strict evaluation metric. Even though only one slot at a turn is mispredicted, the joint goal accuracy of this turn is 0. Thus, the joint goal accuracy of a turn takes the value either 1 or 0.

\begin{table*}[]
\caption{Joint goal accuracy of various methods on the test sets of MultiWOZ 2.0 and MultiWOZ 2.1 ($\pm$ denotes the standard deviation). $\dagger$ indicates the results reported in the original papers. $\star$ means the results reproduced by us using the source codes. $\ddagger$ demonstrates a statistically significant improvement to the best baseline at the 0.01 level using a paired two-sided t-test.}
\vspace{-0.1cm}
\label{tab:mainresults}
\setlength{\tabcolsep}{2.5mm}
\begin{tabular}{lcccccc}
\toprule
\multirow{2}{*}{Model} &
  \multirow{2}{*}{Context Encoder} &
  \multirow{2}{*}{Dialogue History} &
  \multirow{2}{*}{Extra Information} &
  \multirow{2}{*}{\begin{tabular}[c]{@{}c@{}}Slot \\ Correlations\end{tabular}} &
  \multicolumn{2}{c}{Joint Goal Accuracy ($\%$)} \\ \cline{6-7} 
           &       &                      &                             &          & \multicolumn{1}{c}{MultiWOZ 2.0}   & MultiWOZ 2.1   \\ \midrule
SST$^\dagger$        & GAT   & Previous Turn        & Schema Graph                & \cmark  & \multicolumn{1}{c}{51.17}          & 55.23          \\
SAS$^\dagger$        & BiGRU & Previous $\mu$ Turns & -                           & \cmark  & \multicolumn{1}{c}{51.03}          & -              \\
CREDIT-RL$^\dagger$  & BiGRU & Previous $\mu$ Turns & -                           & \xmark & \multicolumn{1}{c}{51.68}          & 50.61          \\
STARC$^\dagger$      & BERT  & Full History         & Auxiliary Datasets & \xmark & \multicolumn{1}{c}{-}              & 49.48          \\
CSFN-DST$^\dagger$       & BERT  & No History           & Schema Graph                & \cmark  & \multicolumn{1}{c}{51.57}          & 52.88          \\
SOM-DST$^\dagger$    & BERT  & Previous Turn        & -                           & \xmark & \multicolumn{1}{c}{51.72}          & 53.01          \\
CHAN$^\star$       & BERT  & Full History         & -                           & \xmark & \multicolumn{1}{c}{53.06}          & 53.38          \\
TripPy$^\dagger$     & BERT  & Full History         & Action+Label Map   & \xmark & \multicolumn{1}{c}{-}              & 55.29          \\
NP-DST $^\dagger$    & GPT-2 & Full History         & -                           & \xmark & \multicolumn{1}{c}{44.03}          & -              \\
SimpleTOD$^\star$  & GPT-2 & Full History         & -                           & \xmark & \multicolumn{1}{c}{-}              & 51.89          \\
MinTL-T5$^\dagger$   & T5    & Previous $\mu$ Turns & -                           & \xmark & \multicolumn{1}{c}{52.07}          & 52.52          \\
MinTL-BART$^\dagger$ & BART-Large  & Previous $\mu$ Turns & -                           & \xmark & \multicolumn{1}{c}{52.10}          & 53.62          \\ \midrule
\textbf{STAR}       & BERT  & Full History         & -                           & \cmark  & \multicolumn{1}{c}{\textbf{54.53$\pm$0.21}$^{\ddagger}$} & \textbf{56.36$\pm$0.34}$^{\ddagger}$ \\ \bottomrule
\end{tabular}
\end{table*}

\subsection{Training Details}

We employ the pre-trained BERT-base-uncased model\footnote{\url{https://huggingface.co/transformers/model_doc/bert.html}} as the dialogue context encoder. This model has 12 layers with 768  hidden units and 12 self-attention heads. We also utilize another BERT-base-uncased model as the slot and value encoder. For the slot and value encoder, the weights of the pre-trained BERT model are frozen during training. For the slot-token attention and slot self-attention, we set the number of attention heads to 4. The number of slot self-attention layers (i.e., $L$) is fixed at 6. We treat the context encoder part of our model as an encoder and the remaining part as a decoder. The hidden size of the decoder (i.e., $d$) is set to 768, which is the same as the dimensionality of BERT outputs. The BertAdam \cite{kingma2014adam} is adopted as the optimizer and the warmup proportion is fixed at 0.1. Considering that the encoder is a pre-trained BERT model while the decoder needs to be trained from scratch, we use different learning rates for the two parts. Specifically, the peak learning rate is set to 1e-4 for the decoder and 4e-5 for the encoder. We use a training batch size of 16 and set the dropout \cite{srivastava2014dropout} probability to 0.1. We also exploit the word dropout technique \cite{bowman2015generating} to partially mask the dialogue utterances by replacing some tokens with a special token $[UNK]$. The word dropout rate is set to 0.1. Note that we do not use word dropout on the previous dialogue state, even though it is part of the input. The maximum input sequence length is set to 512. The best model is chosen according to the performance on the validation set. We run the model with different random seeds and report the average results. For MultiWOZ 2.0 and MultiWOZ 2.1, we apply the same hyperparameter settings.

\section{Experimental Results}

\subsection{Baseline Comparison}

The joint goal accuracy of our model and various baselines on the test sets of  MultiWOZ 2.0 and MultiWOZ 2.1 are shown in Table~\ref{tab:mainresults}\footnote{We noted that both CHAN and SimpleTOD ignore the ``{\itshape dontcare}'' slots in the released source codes, which leads to unreasonably high performance. For a fair comparison, we reproduced the results with all ``{\itshape dontcare}'' slots being considered. Our approach can achieve $60\%$ joint goal accuracy if ignoring the ``{\itshape dontcare}'' slots as well.}, in which we also summarize several key differences of these models. As can be seen, our approach consistently outperforms all baselines on both MultiWOZ 2.0 and MultiWOZ 2.1. Compared with the three methods that have taken slot correlations into consideration (i.e., SST, SAS and CSFN-DST), our approach achieves $2.96\%$ and $1.13\%$ absolute performance promotion on MultiWOZ 2.0 and MultiWOZ 2.1, respectively. Our approach also outperforms other baselines by $1.47\%$ and $1.07\%$ separately on the two datasets. From Table~\ref{tab:mainresults}, we observe that SST and TripPy are the best performing baselines. Both methods reach higher than $55\%$ joint goal accuracy on MultiWOZ 2.1. However, SST needs to construct a schema graph by involving some prior knowledge manually. The schema graph is exploited to capture the correlations among slots. Even though SST leverages some prior knowledge, it is still inferior to our approach. This is because the schema graph only considers the relationships among slot names and thus cannot describe the slot correlations completely. It is worth mentioning that SST also shows a deficiency in utilizing dialogue history. SST achieves the best performance when only the previous turn dialogue history is considered \cite{chen2020schema}. TripPy shows the best performance among BERT-based baselines. However, it employs both system actions and a label map as extra supervision. The label map is created according to the labels in the training portion of the dataset. During the testing phase, the label map is leveraged to correct the predictions (e.g., mapping ``{\itshape downtown}'' to ``{\itshape centre}''). The label map is useful, but it may oversmooth some predictions. On the contrary, our model doesn't rely on any extra information and is a fully data-driven approach. Hence, our model is more general and more scalable. Since our model also achieves better performance than SST and TripPy, we can conclude that it is beneficial to take the slot correlations into consideration. The slot self-attention mechanism proposed by us is able to capture the relevance among slots in a better way.

\begin{table}[]
\caption{Performance comparison between TripPy and STAR with and without the label map being used on MultiWOZ 2.1.}
\vspace{-0.1cm}
\label{tab:labelmap}
\begin{tabular}{lcc}
\toprule
Model  & Label Map & Joint Goal Accuracy ($\%$) \\ \midrule
TripPy$-$ & \xmark    & 44.90                      \\
TripPy & \cmark    & 55.29                      \\
\textbf{STAR}   & \xmark    & 56.36                      \\
\textbf{STAR$+$}   & \cmark    & \textbf{56.58}         \\
\bottomrule
\end{tabular}
\vspace{-0.1cm}
\end{table}

We conduct a further comparison between TripPy and our model STAR with and without the label map being leveraged on MultiWOZ 2.1. We denote TripPy as \textbf{TripPy$-$} when the label map is removed and represent STAR as \textbf{STAR$+$} when the label map is involved. The results are reported in Table~\ref{tab:labelmap}. As can be observed, the performance of TripPy degrades dramatically if the label map is ignored. However, the label map doesn't have significant impacts on the performance of our approach. With the label map being considered, our approach only shows slightly better performance. 

\begin{figure}[t!]
	\centering
	\subfigure[{Single Domain}]{
		\includegraphics[width=0.476\columnwidth]{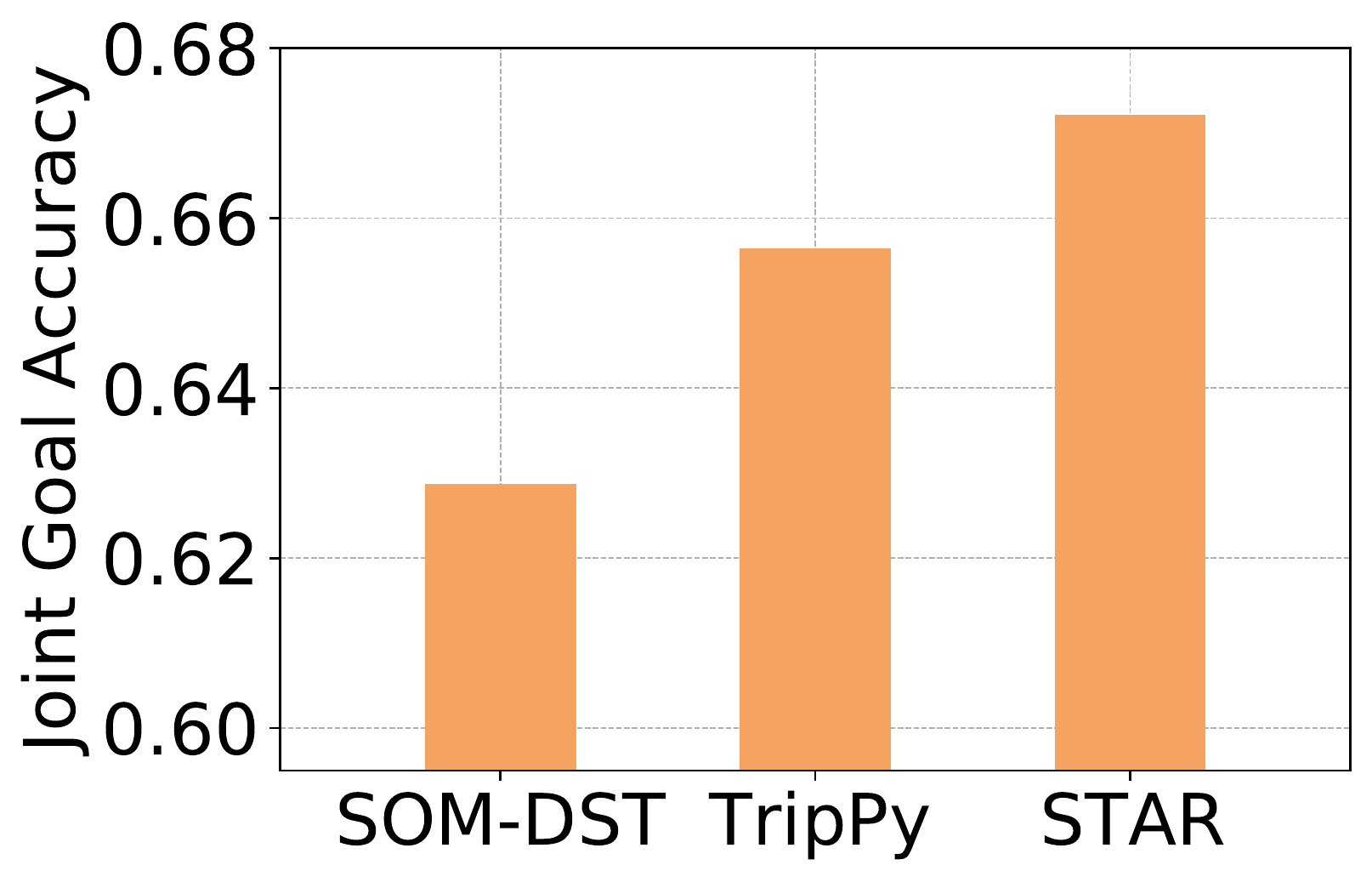}
	} \hfill
	\subfigure[{Multi Domain}]{
		\includegraphics[width=0.476\columnwidth]{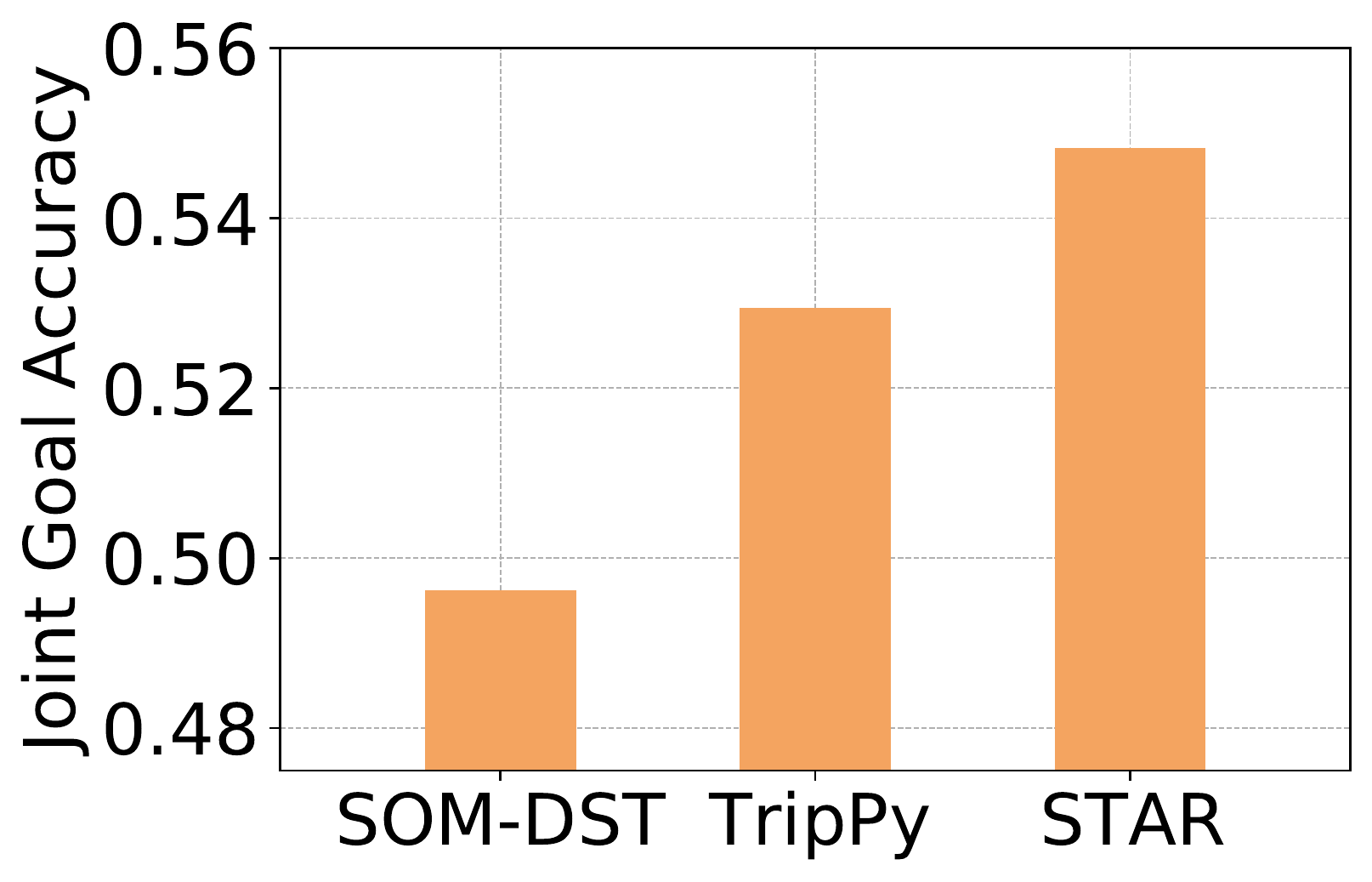}
	}
	\vspace*{-0.4cm}
	
	\caption{Joint goal accuracy of single-domain dialogues and multi-domain dialogues on the test set of MultiWOZ 2.1.}
	\label{fig:singleandmulti}
	\vspace*{-0.25cm}
\end{figure}

\subsection{Single-Domain and Multi-Domain Joint Goal Accuracy}

Considering that a practical dialogue may involve multiple domains or just a single domain, it is useful to explore how our approach performs in each scenario. To this end, we report the joint goal accuracy of single-domain dialogues and multi-domain dialogues on the test set of MultiWOZ 2.1, respectively. The results are shown in Figure~\ref{fig:singleandmulti},  from which we observe that our approach achieves better performance in both scenarios. The results indicate that our approach can capture the correlations among slots both within a single domain and across different domains. From Figure ~\ref{fig:singleandmulti}, we also observe that all the methods demonstrate higher performance in the single-domain scenario. Our approach even reaches about $67\%$ joint goal accuracy. While the performance of all methods in the multi-domain scenario slightly goes down, compared to the overall joint goal accuracy shown in Table~\ref{tab:mainresults}. Nonetheless, our approach still achieves about $55\%$ joint goal accuracy. The results further confirm the effectiveness of our approach.

\subsection{Domain-Specific Joint Goal Accuracy and Slot-Specific Accuracy}

In this part, we first investigate the performance of our model in each domain. The domain-specific joint goal accuracy on MultiWOZ 2.1 is reported in Table~\ref{tab:domainres}, where we compare our approach with CSFN-DST, SOM-DST and TripPy. The domain-specific accuracy is calculated on a subset of the predicted dialogue state. The subset consists of all the slots specific to a domain. In addition, only the domain-active dialogues are considered for each domain. As shown in Table~\ref{tab:domainres}, our approach consistently outperforms CSFN-DST and SOM-DST in all domains. Our approach also outperforms TripPy in three domains. Although TripPy demonstrates better performance in the ``{\itshape attraction}'' domain and ``{\itshape restaurant}'' domain, it shows the worst performance in the ``{\itshape taxi}'' domain. As analyzed in \cite{kim2020somdst}, the ``{\itshape taxi}'' domain is the most challenging one. This domain also has the least number of training dialogues (refer to Table~\ref{tab:datastat}). Owing to the strong capability of modeling slot correlations, our approach achieves much better performance in this challenging domain.

\begin{table}[]
\caption{Domain-specific accuracy ($\%$) on MultiWOZ 2.1.}
\vspace{-0.1cm}
\label{tab:domainres}
\begin{tabular}{lcccc}
\toprule
Domain     & CSFN-DST & SOM-DST & TripPy         & \textbf{STAR}           \\ \midrule
Attraction & 64.78    & 69.83   & \textbf{73.37} & 70.95          \\
Hotel      & 46.29    & 49.53   & 50.21          & \textbf{52.99} \\
Restaurant & 64.64    & 65.72   & \textbf{70.47} & 69.17          \\
Taxi       & 47.35    & 59.96   & 37.54          & \textbf{66.67} \\
Train      & 69.79    & 70.36   & 72.51          & \textbf{75.10} \\ \bottomrule
\end{tabular}
\end{table}

\begin{figure}[t]
  \centering
  \includegraphics[width=0.478\textwidth]{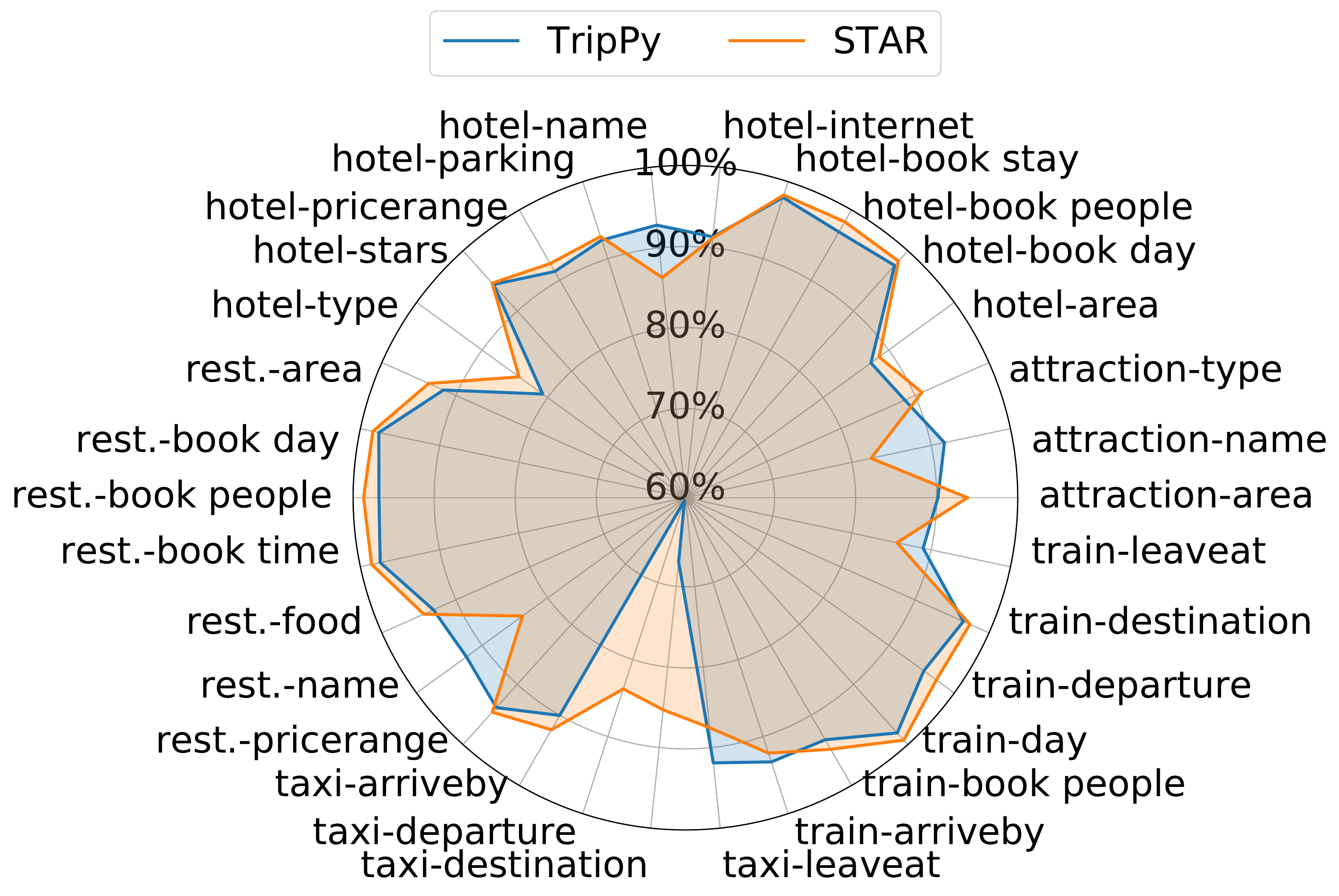}
  \caption{Slot-specific accuracy on MultiWOZ 2.1. The domain ``{\itshape restaurant}'' is represented as ``{\itshape rest.}'' for short.}
  \label{fig:slotacc}
   \vspace*{-0.3cm}
\end{figure}

We then illustrate the slot-specific accuracy of our approach and TripPy in Figure~\ref{fig:slotacc}. The corresponding exact numbers are shown in Table~\ref{tab:slotaccnumber} in Appendix~\ref{sec:app}. The slot-specific accuracy measures the accuracy of each individual slot. Note that the slot-specific accuracy is calculated using only the dialogues that involve the domain the slot belongs to. From Figure~\ref{fig:slotacc}, we observe that both methods demonstrate high performance for most slots. However, TripPy shows relatively poor performance for slot ``{\itshape taxi-departure}'' and slot ``{\itshape taxi-destination}''. The results are consistent with the domain-specific accuracy and explain why TripPy fails in the ``{\itshape taxi}'' domain. From Figure~\ref{fig:slotacc}, we also observe that our approach is inferior to TripPy in the {\itshape name}-related slots (i.e., ``{\itshape attraction-name}'', ``{\itshape hotel-name}'' and ``{\itshape restaurant-name}'') and {\itshape leaveat}-related slots (i.e., ``{\itshape taxi-leaveat}'' and ``{\itshape train-leaveat}''). The values of these slots are usually informed by the users explicitly. Since TripPy leverages copy mechanisms to extract values directly, it seems to be more appropriate for these slots. This observation inspires us that it should be beneficial to extend our model by incorporating the copy mechanism, which we leave as our future work.

\begin{figure}[t]
  \centering
  \includegraphics[width=0.475\textwidth]{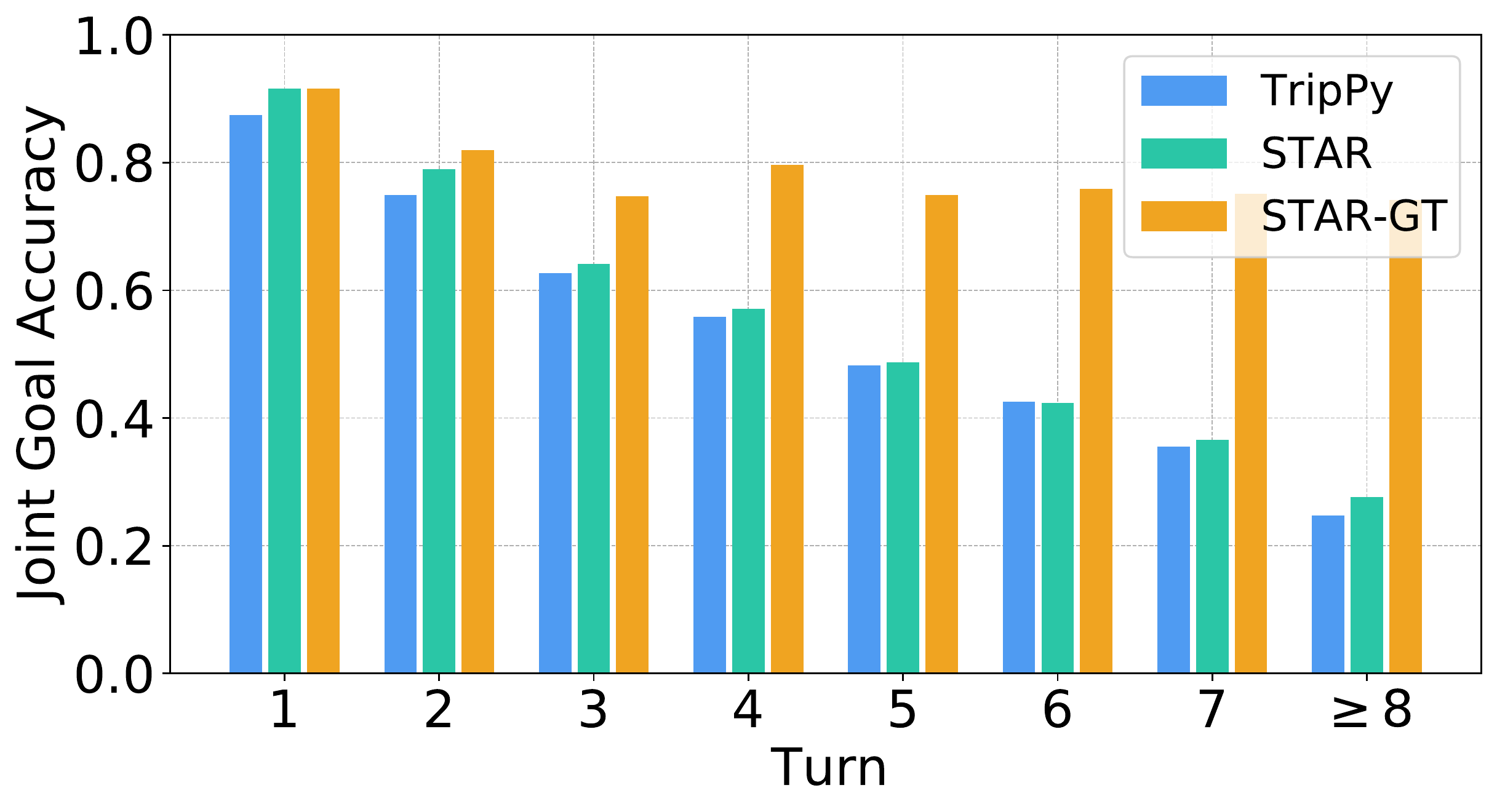}
  \vspace*{-0.7cm}
  \caption{Per-turn joint goal accuracy on MultiWOZ 2.1.}
  \label{fig:turnacc}
  \vspace{-0.3cm}
\end{figure}

\subsection{Per-Turn Joint Goal Accuracy}

Given that practical dialogues have a different number of turns and longer dialogues tend to be more challenging, we further analyze the relationship between the depth of conversation and accuracy of our model. The per-turn accuracy on MultiWOZ 2.1 is shown in Figure~\ref{fig:turnacc}. For comparison, we also include the results of TripPy and \textbf{STAR-GT}. STAR-GT means the groundtruth previous dialogue state is used as the input at each turn. Figure~\ref{fig:turnacc} shows that the accuracy of both TripPy and STAR decreases with the increasing of dialogue turns. In contrast, the performance of STAR-GT is relatively stable. This is because errors occurred in early turns will be accumulated to later turns in practice. However, when the groundtruth previous dialogue state is used, there is no error accumulation.

\begin{figure}
\minipage{0.233\textwidth}
    \centering
    \includegraphics[width = 0.98\columnwidth]{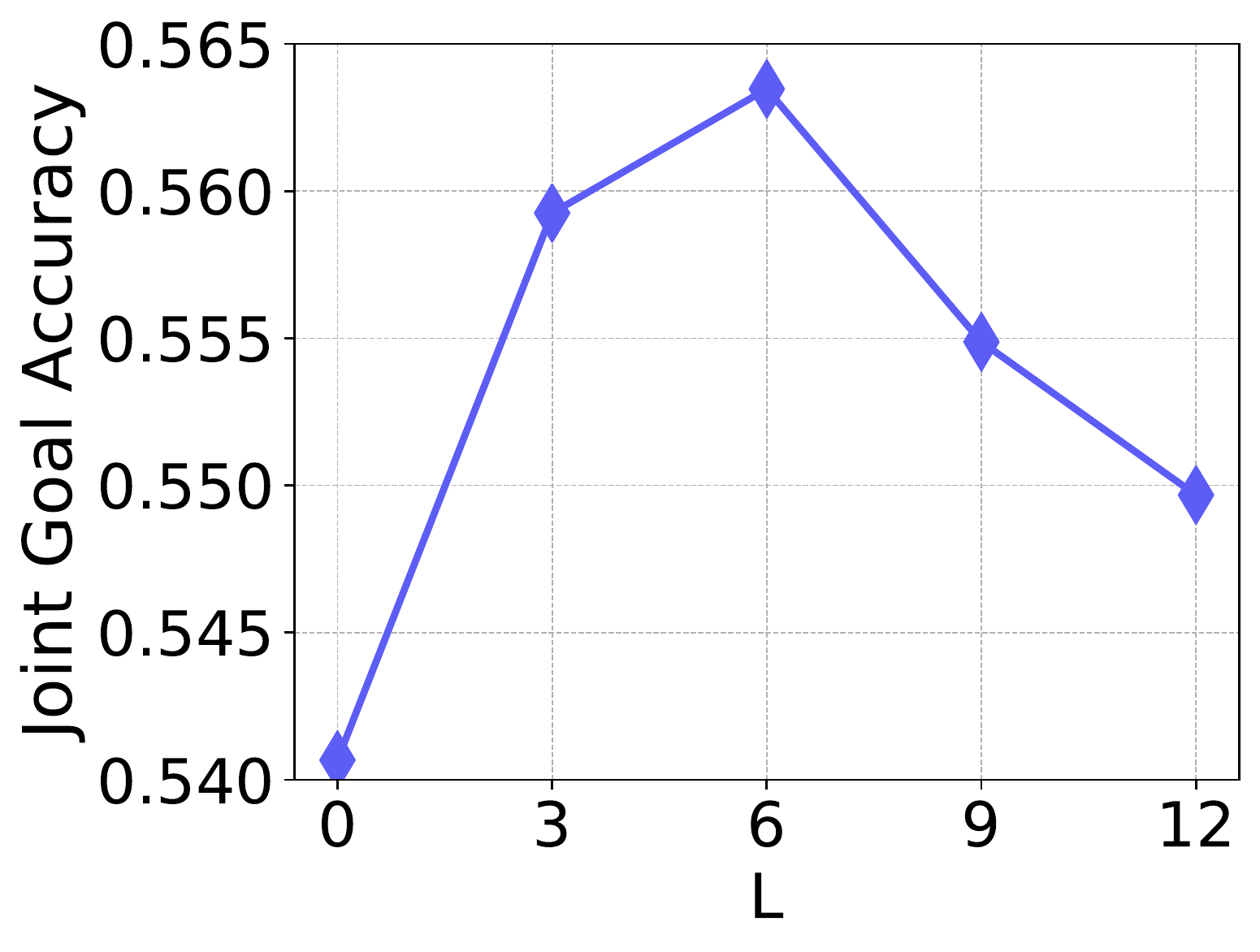}
    \vspace{-0.2cm}
    \caption{Effects of number of slot self-attention layers.}
    \label{fig:layer}
\endminipage \hfill
\minipage{0.233\textwidth}
    \includegraphics[width = 0.98\columnwidth]{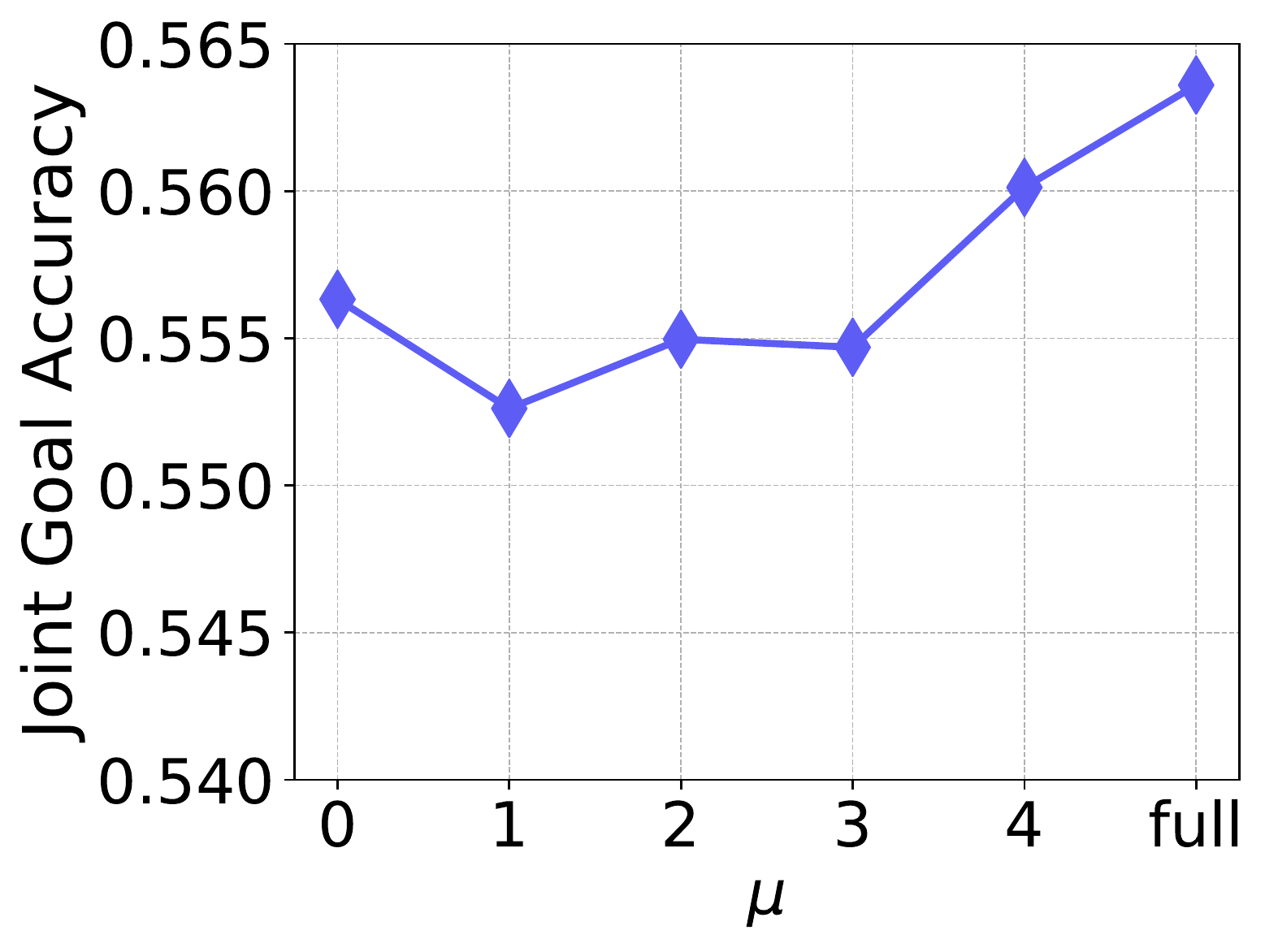}
    \vspace{-0.2cm}
    \caption{Effects of number of previous dialogue turns.}
    \label{fig:history}
\endminipage
 \vspace{-0.18cm}
\end{figure}

\subsection{Effects of Number of Slot Self-Attention Layers}

We vary the number of slot self-attention layers (i.e., $L$) in the range of $\{0, 3, 6, 9, 12\}$ to study its impacts on the performance of our model. The results on MultiWOZ 2.1 are illustrated in Figure~\ref{fig:layer}, from which we observe that our model achieves the best performance when $L$ is set to 6. The performance degrades while $L$ goes larger. This may be caused by overfitting. Figure~\ref{fig:layer} also shows that when there is no slot self-attention (i.e., $L=0$), the joint goal accuracy decreases to around $54\%$. Note that when $L=0$, our model doesn't learn the slot correlations any more. Hence, we conclude that it is essential to take the dependencies among slots into consideration.

\subsection{Effects of Number of Previous Dialogue Turns}

To evaluate the effects of the number of previous dialogue turns (i.e., $\mu$), we vary $\mu$ in the range of $\{0, 1, 2, 3, 4, full\}$, where $full$ means all the previous dialogue turns are considered. The results on MultiWOZ 2.1 are shown in Figure~\ref{fig:history}. As can be seen, when full dialogue history is leveraged, our model demonstrates the best performance. When no dialogue history is employed, our model also reaches higher than $55.5\%$ joint goal accuracy. However, when $\mu$ is set to 1, 2 and 3, the performance degrades slightly. This is probably because the incomplete history leads to confusing information and makes it more challenging to extract the appropriate slot values.

\section{Conclusion}
In this paper, we have presented a novel DST approach STAR to modeling the correlations among slots. STAR first employs a slot-token attention to retrieve slot-specific information for each slot from the dialogue context. It then leverages a stacked slot self-attention to learn the dependencies among slots. STAR is a fully data-driven approach. It does not ask for any human efforts or prior knowledge when measuring the slot correlations. In addition, STAR considers both slot names and their corresponding values to model the slot correlations more precisely. To evaluate the performance of STAR, we have conducted a comprehensive set of experiments on two large multi-domain task-oriented dialogue datasets MultiWOZ 2.0 and MultiWOZ 2.1. The results show that STAR achieves state-of-the-art performance on both datasets. For future work, we intend to incorporate the copy mechanism into STAR to enhance its performance further.


\bibliographystyle{ACM-Reference-Format}
\bibliography{STAR}

\appendix

\section{Slot-Specific Accuracy} 
\label{sec:app}

\begin{table}[h]
\caption{Slot-specific accuracy ($\%$) computed via considering all dialogues in the test set of MultiWOZ 2.1 and only the dialogues that contain the domain the slot belongs to (in gray).}
\label{tab:slotaccnumber}
\small
\setlength{\tabcolsep}{0.44mm}
\begin{tabular}{lccc>{\columncolor[gray]{0.9}}c
>{\columncolor[gray]{0.9}}c
>{\columncolor[gray]{0.9}}c}
\hline
Slot                   & SOM-DST        & TripPy         & STAR           & SOM-DST & TripPy         & STAR           \\ \hline
attraction-area        & 97.08          & 96.64          & \textbf{97.76} & 92.02   & 90.15          & \textbf{93.80} \\
attraction-name        & 94.46          & \textbf{97.16} & 94.20          & 83.44   & \textbf{91.67} & 82.42          \\
attraction-type        & 96.42          & 96.16          & \textbf{96.68} & 89.39   & 88.63          & \textbf{90.94} \\
hotel-area             & \textbf{95.93} & 95.53          & 95.81          & 88.40   & 87.29          & \textbf{88.54} \\
hotel-book day         & 99.14          & 99.14          & \textbf{99.40} & 97.56   & 97.56          & \textbf{98.29} \\
hotel-book people      & 99.14          & 98.90          & \textbf{99.41} & 97.56   & 96.87          & \textbf{98.32} \\
hotel-book stay        & 99.21          & 99.25          & \textbf{99.40} & 97.75   & 97.95          & \textbf{98.28} \\
hotel-internet         & 96.76          & \textbf{96.97} & 96.93          & 90.76   & 91.38          & \textbf{91.42} \\
hotel-name             & 94.34          & \textbf{97.38} & 95.20          & 84.19   & \textbf{92.82} & 86.31          \\
hotel-parking          & 96.95          & 97.34          & \textbf{97.51} & 91.34   & 92.42          & \textbf{92.89} \\
hotel-pricerange       & 97.00          & 96.86          & \textbf{97.15} & 91.61   & 91.23          & \textbf{92.37} \\
hotel-stars            & 97.87          & 98.03          & \textbf{98.08} & 93.93   & 94.40          & \textbf{94.63} \\
hotel-type             & 93.68          & 93.19          & \textbf{94.51} & 82.14   & 80.76          & \textbf{84.37} \\
restaurant-area        & 96.81          & 96.56          & \textbf{97.39} & 92.60   & 91.67          & \textbf{93.67} \\
restaurant-book day    & 99.36          & 99.07          & \textbf{99.39} & 98.40   & 97.67          & \textbf{98.43} \\
restaurant-book people & 99.13          & 98.73          & \textbf{99.38} & 97.81   & 96.81          & \textbf{98.71} \\
restaurant-book time   & 98.82          & 99.00          & \textbf{99.43} & 97.01   & 97.50          & \textbf{98.59} \\
restaurant-food        & 97.22          & 97.20          & \textbf{97.78} & 93.06   & 93.03          & \textbf{94.34} \\
restaurant-name        & 92.78          & \textbf{96.92} & 93.68          & 82.13   & \textbf{92.40} & 83.84          \\
restaurant-pricerange  & 97.34          & 97.34          & \textbf{97.51} & 93.89   & 93.86          & \textbf{94.61} \\
taxi-arriveby          & 99.09          & 99.06          & \textbf{99.28} & 90.65   & 90.03          & \textbf{92.07} \\
taxi-departure         & 97.75          & 96.31          & \textbf{98.49} & 76.63   & 59.35          & \textbf{83.81} \\
taxi-destination       & 98.26          & 96.89          & \textbf{98.61} & 83.17   & 66.98          & \textbf{85.32} \\
taxi-leaveat           & 99.05          & \textbf{99.26} & 98.90          & 89.40   & \textbf{91.90} & 87.54          \\
train-arriveby         & 96.65          & \textbf{97.30} & 96.85          & 91.75   & \textbf{93.28} & 92.14          \\
train-book people      & 97.41          & 97.39          & \textbf{97.95} & 93.52   & 93.49          & \textbf{94.85} \\
train-day              & 98.72          & 99.17          & \textbf{99.63} & 96.98   & 98.06          & \textbf{99.25} \\
train-departure        & 98.26          & 98.01          & \textbf{98.83} & 95.79   & 95.35          & \textbf{97.26} \\
train-destination      & \textbf{98.75} & 98.53          & \textbf{98.75} & 97.11   & 96.54          & \textbf{97.42} \\
train-leaveat          & 94.30          & \textbf{95.57} & 94.28          & 85.79   & \textbf{88.98} & 85.69          \\ \hline
\end{tabular}
\end{table}

\begin{table*}[]
\caption{Dialogue state outputs of TripPy and our model STAR on a dialogue (dialogue\_id: MUL2686.json) from the test set of MultiWOZ 2.1. For simplicity, only the turn-active dialogue states are presented. The false predictions are shown in blue.}
\label{tab:casestudy}
\begin{tabular}{l}
\toprule
\textbf{User:} Hi! Can you help me find some places to go on the \colorbox{red!40}{south} side of the city? I am super bored! \\
\textbf{GroundTruth:} {\itshape attraction-area=south} \\
\textbf{TripPy:} {\itshape attraction-area=south} \\
\textbf{STAR:} {\itshape attraction-area=south} \\ 
\midrule
\textbf{Sys:} I am happy to help. There are several types of places to go in the south, parks, theatre, cinema, museum, nightclub. \\
\hspace{1.76em}What sounds good to you? \\
\textbf{User:} I \colorbox{red!40}{do not have a preference}. Could you make a recommendation? \\
\textbf{GroundTruth:} {\itshape attraction-type=dontcare} \\
\textbf{TripPy:} {\itshape \colorbox{blue!40}{attraction-type=none}} \\
\textbf{STAR:} {\itshape attraction-type=dontcare} \\ \midrule
\textbf{Sys:} How about a nightclub or cinema? There's also a great museum. \\
\textbf{User:} Anything is fine, whatever is most popular. \\
\textbf{TripPy:} {\itshape \colorbox{blue!40}{attraction-type=dontcare}} \\
 \midrule
\textbf{Sys:} Sure, \colorbox{red!40}{cineworld cinema} will be great. What information do you need about this attraction? \\
\textbf{User:} Can I get the address, entrance fee and attraction type please? \\
\textbf{GroundTruth:} {\itshape attraction-name=cineworld cinema} \\
\textbf{TripPy:} {\itshape attraction-name=cineworld cinema} \\
\textbf{STAR:} {\itshape attraction-name=cineworld cinema} \\ \midrule
\textbf{Sys:} The address is cambridge leisure park, clifton way. It is a cinema. I do not know the entrance fee, unfortunately. \\
\textbf{User:} I am also wanting some information on a hotel called the \colorbox{red!40}{kirkwood house}. Can you tell me about them? \\
\textbf{GroundTruth:} {\itshape hotel-name=kirkwood house} \\
\textbf{TripPy:} {\itshape hotel-name=kirkwood house} \\
\textbf{STAR:} {\itshape hotel-name=kirkwood house} \\ \midrule
\textbf{Sys:} That is a guest house found in the north. It is in the moderate price range and is rated 4 stars. Do you need their contact details? \\
\textbf{User:} I need a booking there for \colorbox{red!40}{5 people} staying \colorbox{red!40}{3 nights} from \colorbox{red!40}{Monday}. \\
\textbf{GroundTruth:} {\itshape hotel-book people=5; hotel-book stay=3; hotel-book day=monday} \\
\textbf{TripPy:} {\itshape hotel-book people=5; hotel-book stay=3; hotel-book day=monday} \\
\textbf{STAR:} {\itshape hotel-book people=5; hotel-book stay=3; hotel-book day=monday} \\ \midrule
\textbf{Sys:} Excellent. You are all booked. Reference number is j0vgu1ux. Can I help with anything else? \\
\textbf{User:} Yes. I need a taxi to \colorbox{red!40}{commute between the 2 places}. I would like to \colorbox{red!40}{leave the park} by \colorbox{red!40}{01:15}. \\
\textbf{GroundTruth}: {\itshape taxi-leaveat=01:15; taxi-departure=cineworld cinema; taxi-destination=kirkwood house} \\
\textbf{TripPy:} {\itshape \colorbox{blue!40}{taxi-leaveat=01}; \colorbox{blue!40}{taxi-departure=kirkwood house}; \colorbox{blue!40}{taxi-destination=cineworld cinema}} \\
\textbf{STAR:} {\itshape taxi-leaveat=01:15; taxi-departure=cineworld cinema; taxi-destination=kirkwood house} \\ 
\bottomrule
\end{tabular}
\end{table*}

Traditionally, the slot-specific accuracy for each slot is computed based on all the dialogues in the test set \cite{shan-etal-2020-contextual, lee2019sumbt, hu2020sas, kim2020somdst, heck2020trippy}. However, for a given slot, there are many dialogues that do not contain the domain the slot belongs to. In these dialogues, the value of the slot is always \verb|none|. Due to this, even if a model predicts the values of a slot for all dialogues as \verb|none|, it can still achieve a relatively high slot-specific accuracy. To overcome this limitation, we propose calculating the slot-specific accuracy using only the dialogues that involve the domain the slot belongs to. The detailed results are shown in Table~\ref{tab:slotaccnumber} (in gray). For comparison, we also include the results computed based on all dialogues. As can be seen, no matter which method is adopted to calculate the slot-specific accuracy, our model is able to achieve better performance for most slots. Table~\ref{tab:slotaccnumber} also shows that if the traditional method is adopted, all three models demonstrate higher than $90\%$ slot-specific accuracy for each slot. Besides, there are only subtle differences in the slot-specific accuracy of the three models. While the slot-specific accuracy computed using our proposed method is more discriminative.

\section{Case Study}

Table~\ref{tab:casestudy} shows the predicted dialogue states of our model and TripPy on a dialogue from the test set of MultiWOZ 2.1. As can be seen, our model correctly predicts all slot values at each turn. However, TripPy fails to predict the value of slot ``{\itshape attraction-type}'' at turn 2 and delays the prediction to turn 3. At the last turn, TripPy predicts the value of slot ``{\itshape taxi-leaveat}'' as ``{\itshape 01}'' rather than ``{\itshape 01:15}'', albeit this information is explicitly contained in the user utterance. For slot ``{\itshape taxi-departure}'' and slot ``{\itshape taxi-destination}'', since the user provides the corresponding information indirectly, it is challenging to deduce their valid values. TripPy falsely predicts the destination as the departure and vice versa.

\end{document}